\documentclass{article} 
\usepackage{iclr2026/shmiclr2026_conference,times}
\iclrfinalcopy

\usepackage{amsmath,amsfonts,bm}

\def\eqref#1{equation~\ref{#1}}

\def\1{\bm{1}}

\DeclareMathAlphabet{\mathsfit}{\encodingdefault}{\sfdefault}{m}{sl}
\SetMathAlphabet{\mathsfit}{bold}{\encodingdefault}{\sfdefault}{bx}{n}

\usepackage{hyperref}
\usepackage{url}
\usepackage{booktabs}
\usepackage{array}
\usepackage{xcolor}
\usepackage{colortbl}
\usepackage{multirow}
\usepackage{algorithm}
\usepackage{graphicx}
\usepackage[capitalize,noabbrev,nameinlink]{cleveref}

\usepackage[font=small,labelfont=bf]{caption}
\usepackage{enumitem}

\definecolor{professionalblue}{rgb}{0.2, 0.5, 0.8}   
\definecolor{mediumblue}{rgb}{0.3, 0.6, 0.85}        
\definecolor{steelblue}{rgb}{0.27, 0.51, 0.71}      
\definecolor{darkgray}{rgb}{0.4, 0.4, 0.4}       

\hypersetup{
    colorlinks=true,
    linkcolor=professionalblue,  
    filecolor=darkgray,         
    urlcolor=steelblue,        
    citecolor=mediumblue
    }

\title{Rewiring Experts on the Fly: \\Continuous Rerouting for Better Online Adaptation in Mixture-of-Expert models}

\author{
 Guinan Su\textsuperscript{1*}\quad
 Yanwu Yang\textsuperscript{4*}\quad
 Li Shen\textsuperscript{5}\quad
 Lu Yin\textsuperscript{6}\quad
 Shiwei Liu\textsuperscript{1,2,3}\quad
 Jonas Geiping\textsuperscript{1,2,3}
 \\[0.5em]
 \small
 \textsuperscript{1}Max Planck Institute for Intelligent Systems\quad
 \textsuperscript{2}ELLIS Institute Tübingen\quad
 \textsuperscript{3}Tübingen AI Center\\
 \textsuperscript{4}University of Tübingen\quad
 \textsuperscript{5}Sun Yat-sen University\quad
 \textsuperscript{6}University of Surrey
}

\usepackage{caption}
\begin{document}

\maketitle
\let\thefootnote\relax
\footnotetext{* Equal contribution.}
 \footnotetext{\texttt{guinan.su@tuebingen.mpg.de}}

\vspace{-0.4cm}
\begin{abstract}

Mixture-of-Experts (MoE) models achieve efficient scaling through sparse expert activation, but often suffer from suboptimal routing decisions 
due to distribution shifts in deployment. While existing test-time adaptation methods could potentially address these issues, they primarily focus on dense models and require access to external data, limiting their practical applicability to MoE architectures. However, we find that, instead of relying on reference data, we can optimize MoE expert selection on-the-fly based only on input context. As such, we propose \textit{a data-free, online test-time framework} that continuously adapts MoE routing decisions during text generation without external supervision or data. Our method cycles between two phases: During the prefill stage, and later in regular intervals, we optimize the routing decisions of the model using self-supervision based on the already generated sequence. Then, we generate text as normal, maintaining the modified router until the next adaption.
%
%
%
%
%
%
We implement this through lightweight additive vectors that only update router logits in selected layers, maintaining computational efficiency while preventing over-adaptation. The experimental results show consistent performance gains on challenging reasoning tasks while maintaining robustness to context shifts. For example, our method achieves a 5.5\% improvement on HumanEval with OLMoE. Furthermore, owing to its plug-and-play property, our method naturally complements existing test-time scaling techniques, e.g., achieving 6\% average gains when incorporated with self-consistency on DeepSeek-V2-Lite.




\end{abstract}



\vspace{-0.5cm}
\section{Introduction}
Mixture-of-Experts (MoE) models \citep{shazeer2017outrageously, zhou2022mixture, jiang2024mixtral, dai2024deepseekmoe, liu2024deepseek, team2025qwen3, muennighoff2024olmoe} provide an effective approach to scaling model capacity while maintaining computational efficiency by partitioning parameters into specialized experts and selectively activating subsets through routing mechanisms \citep{lepikhin2020gshard, fedus2022switch, dai2024deepseekmoe, muennighoff2024olmoe}. This functionality enables dynamic expert selection for diverse queries and creating inherently general-purpose systems that can store much more functionality and information than is used in every forward pass. However, despite their impressive capabilities, MoE models still face challenges when deployed in real-world environments \citep{akyurek2024surprising, li2025c3po}, as the expression of their capability hinges on the quality of their \textit{routing decisions}, the activations of small linear layers that determine which parts of the model are activated.

\looseness -1 Why is routing hard? While the full MoE may store sufficient functionality to solve a particularly challenging query, this capacity is gated behind its routing decisions in each MoE layer, which in turn depend on the residual stream of the model, and so on earlier routing decisions. Routing decisions are only linear functions of the current hidden state that need to approximate the anticipated utility of activating a certain expert. 
A particular issue with this non-robustness of routing is that during standard inference \textit{there is no mechanism to reinforce the routing to a particularly successful part of the model}, or to reduce routing to parts that did not contribute meaningful signal to the generated text. 

This makes routing optimization a question of model adaptation, reminiscent of neuroplasticity in humans -- who continuously optimize routing and neuronal connections in the brain through adaptation and self-regulation, for example when continuing to practice a certain task. Yet, in MoEs, suboptimal expert selection leads to routing inefficiencies, creating a critical bottleneck in overall performance \citep{shi2024unchosen,li2025c3po}.
And, while there has been a significant amount of work to improve the capabilities of MoEs in general, such as through test-time scaling, it is less clear how to best adapt MoEs to different tasks at inference. While standard approaches, such as in-context learning with task-specific demonstrations \citep{wei2022chain,madaan2023self} or parallel generation strategies that produce multiple candidates and aggregate them \citep{wang2022self,brown2024large}, do adapt the model overall, they influence routing only implicitly. 

Conceptually, adaptation should be solvable by \textit{test-time training}, but existing approaches \citep{hardt2305test, hubotter2024efficiently} focus on a classical prediction perspective, which retrieves relevant data from training sets or knowledge bases during inference to fine-tune models for dynamic scenarios before the model is being used. \Citet{li2025c3po} attempt to address router optimization through this perspective of test-time training, and, as such, use retrieval of "successful neighbors" from reference sets based on the first prompt in each context. However, this approach requires access to external reference data during deployment, incurs retrieval overhead, and risks failures in retrieval due to short prompts. Moreover the approach is static, as modern models execute test-time scaling, they generate long chains-of-thought, that should be taken into account when optimizing routing.

\begin{figure}[t] 
    \centering
    \includegraphics[width=0.95\textwidth ]{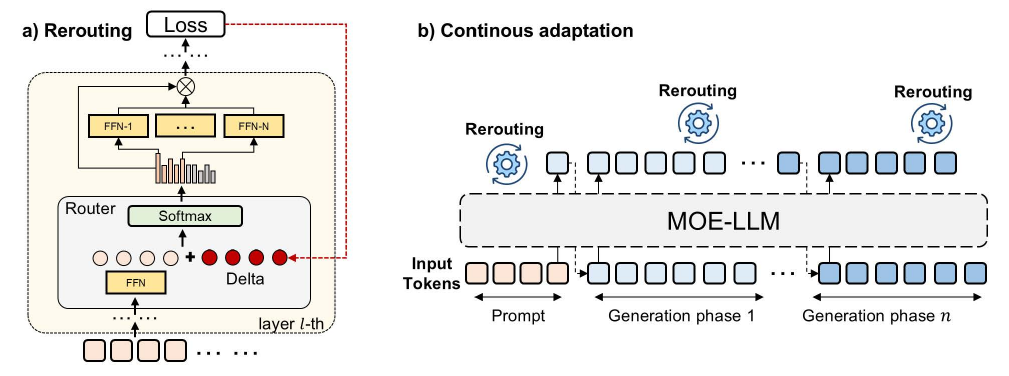}
    \caption{\textbf{Test-time rerouting framework for MoE models.} \textbf{(a) Rerouting mechanism}: lightweight additive vectors (Delta) update router logits in selected high-confidence layers using self-supervised loss from existing context. \textbf{(b) Continuous adaptation}: alternating between optimization phases that adapt routing decisions and generation phases that maintain adapted routing until the next optimization cycle.}
    \label{fig:main}
    \vspace{-0.6cm}
\end{figure}

In this regard, we propose a simple yet effective \emph{data-free} test-time rerouting framework for MoE models as shown in Figure \ref{fig:main} that treats each input prompt as a self-supervised learning opportunity, enabling dynamic rerouting of pretrained MoE models for individual prompts during inference. The framework alternates in two phases: (1) \textbf{In-Context Routing Optimization}, where we regard the current context itself as a training sample and execute optimization steps to minimize cross-entropy loss on the current context with regard to routing logits; and (2) \textbf{Steered Generation}, where we generate text normally, steering routers with updates computed in the previous phase.
This creates a dynamic feedback loop where \emph{the model continuously refines its understanding of task requirements} based on its own generation progress, enabling increasingly informed expert routing decisions as generation proceeds.  To maintain computational efficiency, we implement this progressive optimization through \textbf{lightweight additive parameter vectors that update only the model's router logits of selected MoE layers}, further reducing computational overhead and preventing over-adaptation.
Overall, our main contributions are:
\begin{itemize}[itemsep=0.0pt,topsep=0pt,leftmargin=*]
\item We propose a test-time rerouting framework specifically designed for MoE models, operating completely without external data dependencies or expensive retrieval mechanisms, using only backpropagation within the current context.
\item We introduce a lightweight parameter update mechanism that selectively optimizes router logits through additive vectors in high-confidence layers,
enabling efficient expert selection adaptation during inference using steering.
\item We validate our method's effectiveness through extensive experiments, demonstrating that our approach significantly improves MoE model performance on complex reasoning tasks, maintains robustness to context shifts in multi-turn conversations, and seamlessly integrates with existing test-time techniques, establishing a practical and versatile solution for deploying MoE models in real-world applications.

\end{itemize}
\looseness -1 Overall, we argue that routing changes are a compelling mechanism with which to adapt MoE models on the fly. The changes are lightweight, quick to compute and to apply (even in multi-user settings), and remain effective across context shifts, allowing MoEs a novel degree of plasticity that allows adaptation to task changes during deployment, paving the way toward practical continual self-regulation of MoE models during use.


\section{Related Work}
\textbf{Mixture-Of-Experts (MoE).}
Sparsely activated Mixture-of-Expert models (MoE) are an efficient method for scaling model size by replacing the feed-forward networks (FFN) of transformer architectures with multiple experts (each its own FFN) and a gating function. This architecture dynamically activates different experts for each input token rather than utilizing all parameters \citep{shazeer2017outrageously,lepikhin2020gshard,fedus2022switch}. Recent MoE-based LLMs, including OLMoE \citep{muennighoff2024olmoe}, DeepSeek \citep{dai2024deepseekmoe,liu2024deepseek}, and Qwen2.5 \citep{team2025qwen3}, adopt top-k expert routing to reduce the active parameter count during inference. These models achieve competitive performance while maintaining significantly lower memory costs.

\textbf{Test-Time Scaling.}
Test-Time Scaling enhances LLM capabilities by allocating additional computational resources during inference. These approaches fall into two categories \citep{welleck2024decoding}: Parallel generation, including self-consistency evaluations via multiple candidate responses \citep{wang2022self}, best-of-N sampling \citep{brown2024large}, and Monte Carlo Tree Search \citep{zhou2023language,xie2024monte}, and sequential generation such as extending outputs through chain-of-thought reasoning \citep{wei2022chain, madaan2023self}.

\textbf{Test-Time Training (TTT)}. TTT offers an alternative scaling approach. While successful in computer vision \citep{wang2020tent, sun2020test, sun2024learning, gandelsman2022test, osowiechi2023tttflow}, recent works have extended TTT to language models through fine-tuning on retrieved neighbors \citep{hardt2305test} or optimized data selection algorithms like SIFT \citep{hubotter2024efficiently}. Test-Time Reinforcement Learning (TTRL) \citep{zuo2025ttrl} uses majority voting as reward signals. When applied to Mixture-of-Experts (MoE) models, recent work has explored expert-level interventions for behavior control, enabling precise modifications through targeted expert manipulation \citep{dahlke2025mixture, wang2025two}. Most relevant to our approach, \citet{li2025c3po} optimizes expert routing in MoE models using "successful neighbors" from reference sets. However, these methods assume accessible training data during deployment and introduce significant retrieval overhead, limiting real-world practicality.

\section{Methodology}
This section presents our data-free test-time rerouting framework for MoE models that optimizes expert routing during inference without external information. After reviewing MoE fundamentals (Section \ref{sec:moe-recap}), we introduce three key components: (1) Router Logits Modification (Section \ref{sec:router-logits}) for layer-specific expert selection steering, (2) Dynamic Layer Selection (Section \ref{sec:layer-selection}) for selective MoE layer updates based on confidence scores, and (3) Optimization Procedure (Section \ref{sec:optimization}) detailing our two-phase strategy alternating between in-context routing optimization and steered generation.


\subsection{Preliminaries on Mixture-of-Experts}\label{sec:moe-recap}
In Transformer-based Mixture-of-Experts (MoE) models, the conventional Feed-Forward Networks (FFNs) are replaced with MoE layers. Each MoE layer consists of a router $R$ and a set of experts $\{E_i\}_{i=1}^N$. Given an input sequence $\mathbf{x}_{<t} = (x_1, x_2, \ldots, x_{t-1})$, the router assigns each token to a subset of experts for processing.
Given a token's hidden state $\mathbf{h} \in \mathbb{R}^d$ at layer $l$, the router computes logits across all $N$ experts:
$
\mathbf{z}^{(l)} = W_r^{(l)} \mathbf{h}^{(l)}
$
where $W_r^{(l)} \in \mathbb{R}^{N \times d}$ is the router's weight matrix at layer $l$, and $\mathbf{z}^{(l)} \in \mathbb{R}^N$ represents the logits for expert selection.
These logits are then converted to expert selection probabilities:
$
\mathbf{w}^{(l)} = \text{Softmax}(\mathbf{z}^{(l)})
$
where $w_i^{(l)}$ represents the activation probability for expert $E_i$ at layer $l$.
The router applies a routing strategy (e.g., top-$k$ routing) to select active experts. Weights for unselected experts are zeroed and the remaining weights are renormalized to $\hat{\mathbf{w}}^{(l)}$. The final MoE output is:
$
\mathbf{o}^{(l)} = \sum_{i \in \mathcal{A}^{(l)}} \hat{w}_i^{(l)} \cdot E_i(\mathbf{h}^{(l)})
$
where $\mathcal{A}^{(l)} = \{j \mid \hat{w}_j^{(l)} \neq 0\}$ denotes the set of activated experts at layer $l$.



\subsection{Router Logits Modification}\label{sec:router-logits}
We introduce layer-specific adaptation parameters $\{\boldsymbol{\delta}^{(l)}\}_{l=1}^L$ where $\boldsymbol{\delta}^{(l)} \in \mathbb{R}^{N}$ corresponds to the $N$ experts at MoE layer $l$. For a selected layer $l$, we modify the router logits by adding the corresponding layer-specific parameter:
$
\tilde{\mathbf{z}}^{(l)} = \mathbf{z}^{(l)} + \boldsymbol{\delta}^{(l)}
$
where $\tilde{\mathbf{z}}^{(l)} \in \mathbb{R}^N$ represents the modified logits for layer $l$. The expert selection probabilities are then computed as:
\begin{equation}
\tilde{\mathbf{w}}^{(l)} = \text{Softmax}(\tilde{\mathbf{z}}^{(l)}) = \text{Softmax}(\mathbf{z}^{(l)} + \boldsymbol{\delta}^{(l)})
\end{equation}
This modification directly influences the expert selection distribution, allowing the model to adapt routing decisions based on prompt characteristics.

\subsection{Dynamic Layer Selection}\label{sec:layer-selection}
To mitigate computational overhead and prevent over-adaptation, our framework selectively updates router logits of only a subset of MoE layers rather than all layers simultaneously. We hypothesize that layers with higher routing confidence indicate more decisive and task-relevant expert selection, making them more impactful for adaptation. 
We define router confidence at layer $i$ for token $n$ as:
\begin{equation}
C_i^{(n)} = -\frac{1}{k} \sum_{j=1}^{k} \log p_{i,j}^{(n)}
\end{equation}
where $p_{i,j}^{(n)}$ is the probability of the $j$-th top expert at layer $i$ for token $n$, and $k$ is the number of activated experts per layer. Higher confidence values indicate more decisive routing decisions, suggesting these layers play more critical roles for the current task.
To obtain layer-level confidence across the generated sequence, we aggregate token-level confidence:
\begin{equation}
C_i = \frac{1}{N} \sum_{n=1}^{N} C_i^{(n)}
\end{equation}
where $N$ is the number of generated tokens so far.
We implement two layer selection strategies: (1) \textbf{Hard selection} that selects top-$r$ proportion of layers with highest confidence scores: $\mathcal{S}_t = \text{TopK}(\{C_i\}_{i=1}^L, r)$, and (2) \textbf{Soft weighting} that assigns confidence-based weights to control the update strength of $\boldsymbol{\delta}^{(l)}$:
\begin{equation}
w_i = \frac{C_i}{\sum_{j=1}^{L} C_j}
\end{equation}
During gradient updates, $\boldsymbol{\delta}^{(l)}$ is updated with learning rate scaled by $w_l$.




\subsection{Optimization Procedure}\label{sec:optimization}

\textbf{Parameter Initialization:} For each MoE layer $l \in \mathcal{L}$, initialize routing adjustment parameters:
\begin{equation}
\boldsymbol{\delta}^{(l)} = \mathbf{0} \in \mathbb{R}^{N}
\end{equation}

The framework alternates between two phases:

\textbf{Phase 1: In-Context Routing Optimization.} Given current context $\mathbf{x} = (x_1, \ldots, x_t)$, first select layers using $\mathcal{S} = \text{TopK}(\{C^{(l)}\}_{l=1}^L, r)$ or compute soft weights $\{w_l\}_{l=1}^L$. Then perform $n$ optimization steps, where at each step we compute the cross-entropy loss:
\begin{equation}
\mathcal{L}(\{\boldsymbol{\delta}^{(l)}\}_{l=1}^L) = -\sum_{i=1}^{t-1} \log p(x_{i+1} \mid x_{1:i}, \{\boldsymbol{\delta}^{(l)}\}_{l=1}^L)
\end{equation}
and update parameters using optimizer $\mathcal{O}$ (e.g., SGD, Adam):
\begin{equation}
\boldsymbol{\delta}^{(l)} \leftarrow \begin{cases}
\mathcal{O}(\boldsymbol{\delta}^{(l)}, \nabla_{\boldsymbol{\delta}^{(l)}} \mathcal{L}) & \text{if } l \in \mathcal{S} \text{ (hard selection)} \\
\mathcal{O}(\boldsymbol{\delta}^{(l)}, w_l \nabla_{\boldsymbol{\delta}^{(l)}} \mathcal{L}) & \text{if soft weighting} \\
\boldsymbol{\delta}^{(l)} & \text{otherwise}
\end{cases}
\end{equation}

\textbf{Phase 2: Steered Generation.} Generate $m$ tokens using optimized routing parameters $\{\boldsymbol{\delta}^{(l)}\}_{l=1}^L$.

After generating m tokens, return to Phase 1 with extended context including both original prompt and all generated tokens up to the current position and repeat the optimization procedure. The detailed algorithm is provided in \cref{alg:tamoe}.

\section{Experimental Setup} 
\label{exp-settings}
\textbf{Benchmarks}  We evaluate our approach using a diverse set of benchmarks across multiple reasoning domains. For general knowledge assessment, we employ MMLU-redux\citep{gema2024mmlu}, utilizing generation mode rather than multiple-choice format to encourage deeper reasoning before providing final answers. For code generation tasks, we use HumanEval \citep{chen2021evaluating} and MBPP-sanitized \citep{austin2021program}. For mathematical reasoning, we evaluate on GSM8K \citep{cobbe2021training} and MATH500 \citep{lightman2023let}.

\textbf{Baselines} We compare our method with two adaptation techniques: In-Context Learning (ICL) \citep{wei2022chain,madaan2023self} and C3PO \citep{li2025c3po}. Since our method is a data-free sequential generation test-scaling approach, we select In-Context Learning (ICL) as a primary baseline, using 3 and 5 sample pairs respectively, for comparison. We also compare against C3PO, for which we select a reference set of 100 samples for each dataset. Note that parallel generation methods represent an orthogonal approach to ours, and we further discuss the potential integration of our method with such approaches in later sections.

\textbf{Model Selection} We evaluate three MoE LLMs: OLMoE \citep{muennighoff2024olmoe}, Qwen1.5-MoE \citep{team2024qwen2}, and DeepSeek-V2-Lite \citep{liu2024deepseek}. OLMoE employs 16 layers with 64 experts per layer, activating 8 experts per token (6.9B total, 1.3B active parameters). Qwen1.5-MoE-A2.7B, incorporates 4 shared experts alongside 60 routing experts with 4 activated per token  (14.3B total, 2.7B active parameters). DeepSeek-V2-Lite uses 28 layers with 2 shared and 64 routed experts, activating all shared plus 6 routed experts per token (16B total, 2.4B active parameters).

\textbf{Optimization} We optimize the adaptation parameters using the Adam optimizer with a small number of iterations ($T = 5$) to maintain computational efficiency. The optimization hyperparameters are set as follows: learning rate $\eta = 0.05$, weight decay $= 1 \times 10^{-8}$, and epsilon $= 1 \times 10^{-5}$. All adaptation parameters are initialized to zero: $\boldsymbol{\delta}^{(l)} = \mathbf{0}$. We use the soft-weighting strategy for layer selection, and during the generation stage, we periodically re-optimize the parameters every 128 generated tokens using identical hyperparameter settings to ensure continuous refinement throughout the sequence generation process.

\section{Results for MoE Rewiring} 

\subsection{Main results}
\label{res}
We first evaluated the performance of our method on challenging reasoning benchmarks. As shown in Table \ref{tab:model_performance}, our approach consistently improves performance over all baselines across five benchmarks. On the HumanEval task in particular, our method yields gains of 3.6\%, 5.5\%, and 6.7\% on DeepSeek-V2-Lite, OLMoE, and Qwen1.5-MoE, respectively. Moreover, compared to other calibration methods such as few-shot learning and C3PO, our method achieves consistently better results. Notably, it surpasses these baselines without relying on example demonstrations, delivering particularly strong improvements in code generation and mathematical reasoning tasks. Moreover, in contrast to C3PO, which requires 100 reference samples, our data-free method attains superior performance while requiring no additional reference data.

\begin{table}[h]
\centering
\small
\setlength{\tabcolsep}{5pt}
\renewcommand{\arraystretch}{0.8}
\caption{Model Performance Comparison Across Different Benchmarks, comparing to in-context learning (ICL) and C3PO \cite{li2025c3po}. Rewiring shows strong performance even though no external data is used, either as references or as fewshot examples.}
\vspace{-0.2cm}
\label{tab:model_performance}
\begin{tabular}{l*{6}{c}}
\toprule
\textbf{Method} & \textbf{HumanEval} & \textbf{MBPP} & \textbf{GSM8K} & \textbf{MATH500} & \textbf{MMLU} & \textbf{Average} \\
\midrule
\multicolumn{7}{c}{\textbf{DeepSeek-V2-Lite}} \\
\midrule
Baseline & 50.60 & 58.37 & 72.10 & 22.60 & 50.77 & 50.89 \\
ICL (3-shot) & 52.44 & 56.81 & 71.10 & 21.60 & 44.33 & 49.26 \\
ICL (5-shot) & 53.05 & 52.53 & 71.57 & 22.20 & 46.07 & 49.08 \\
C3PO (100-reference) & 47.80 & 59.92 & 68.80 & 18.20 & 45.77 & 48.10 \\
\rowcolor{blue!12}
Rewiring (\textbf{Ours}, 0-shot) & \textbf{54.26} & \textbf{62.65} & \textbf{73.62} & \textbf{25.00} & \textbf{52.40} & \textbf{53.59} \\
\midrule
\multicolumn{7}{c}{\textbf{OLMoE}} \\
\midrule
Baseline & 28.66 & 40.08 & 51.48 & 10.40 & 37.17 & 33.56 \\
ICL (3-shot) & 28.05 & 36.96 & 43.82 & 9.90 & \textbf{44.90} & 32.73 \\
ICL (5-shot) & 21.21 & 39.30 & 44.96 & 9.40 & 43.63 & 31.70 \\
C3PO (100-reference) & 28.05 & 41.25 & 52.08 & \textbf{13.00} & 37.30 & 34.33 \\
\rowcolor{blue!12}
Rewiring (\textbf{Ours}, 0-shot) & \textbf{34.17} & \textbf{42.80} & \textbf{52.99} & 12.40 & 38.30 & \textbf{36.13} \\
\midrule
\multicolumn{7}{c}{\textbf{Qwen1.5-MoE}} \\
\midrule
Baseline & 40.24 & 46.69 & 54.73 & 21.80 & 45.27 & 41.75 \\
ICL (3-shot) & 44.34 & 46.25 & 55.27 & \textbf{22.20} & 45.60 & 42.73 \\
ICL (5-shot) & 43.90 & 44.75 & 52.31 & 19.40 & \textbf{46.03} & 41.28 \\
C3PO (100-reference) & 39.70 & 45.40 & 50.33 & 18.90 & 44.82 & 39.83 \\
\rowcolor{blue!12}
Rewiring (\textbf{Ours}, 0-shot) & \textbf{46.95} & \textbf{47.47} & \textbf{56.00} & 22.00 & 45.87 & \textbf{43.66} \\

\bottomrule
\end{tabular}
\end{table}

\begin{table}[h]
\centering
\small
\setlength{\tabcolsep}{6pt}
\renewcommand{\arraystretch}{0.8}
\caption{Ablation Study: Layer Selection Strategies and Continuous Refinement.}
\label{tab:ablation_results}
\vspace{-0.2cm}
\begin{tabular}{l*{6}{c}}
\toprule
\textbf{Method} & \textbf{HumanEval} & \textbf{MBPP} & \textbf{GSM8K} & \textbf{MATH500} & \textbf{MMLU} & \textbf{Average} \\
\midrule
Baseline & 50.60 & 58.37 & 72.10 & 22.60 & 50.77 & 50.89 \\
\midrule
\multicolumn{7}{c}{\textit{Layer Selection Strategies}} \\
\midrule
Random Selection & 48.78 & 53.31 & 72.50 & 23.40 & 50.87 & 49.77 \\
Reverse Metric & 48.18 & 55.26 & 72.10 & 21.89 & 49.33 & 49.35 \\
Last-five Layers & 50.60 & 55.64 & 71.85 & 25.20 & 50.33 & 50.72 \\
All Layers & 48.17 & 56.42 & 73.09 & 24.60 & 51.10 & 50.68 \\
\midrule
\multicolumn{7}{c}{\textit{Continuous Refinement}} \\
\midrule
W/o Continuous & 54.27 & 59.53 & 73.37 & 23.80 & 51.80 & 52.55 \\
\midrule
\rowcolor{blue!10}
Rewiring (Complete Method) & \textbf{54.27} & \textbf{62.65} & \textbf{73.62} & \textbf{25.00} & \textbf{52.40} & \textbf{53.59} \\
\bottomrule
\end{tabular}
\vspace{-0.6cm}
\end{table}

\subsection{Ablations of Key Components}
\label{ana}
To validate the effectiveness of our design choices for MoE rewiring, we conduct ablation studies on using  DeepSeek-V2-Lite, examining two key components: layer selection strategies and continuous refinement mechanisms. We evaluate performance across five tasks, with results presented in Table~\ref{tab:ablation_results}.

\textbf{Confidence-based layer selection efficiently discovers task-specific layers.} We compare our confidence-based selection with several baselines, including random selection, selecting the last five layers, selecting low-confidence layers (Reverse Metric), and optimizing all layers. As shown in Table~\ref{tab:ablation_results}, our method achieves an average performance of 53.59\%, outperforming random selection (49.77\%), the last-five-layer strategy used in C3PO (50.72\%), and reverse selection targeting low-confidence layers (49.35\%).
This demonstrates that router confidence identifies layers with strong expert selection knowledge. In effect, we use the signal from prior context to ``confirm'' that routing choices were accurate, and that the model should rely on these experts more strongly.

\looseness -1 \textbf{Updating Single Layers is Safer than rerouting the whole model.} 
In addition, from the results we can see that updating all layers simultaneously (50.68\% on average) underperforms our selective approach (53.59\% on average), highlighting the superiority of our method in selecting layers for test-time rerouting. With limited optimization steps available, spreading updates across all parameters dilutes the refinement signal and risks overadaptation that destabilizes pre-trained routing patterns. Concentrating the optimization budget on high-confidence layers maximizes impact by amplifying existing routing strengths rather than attempting comprehensive corrections across the entire network.

\textbf{Continuous Refinement Stabilizes Routing}.
The comparison between continuous (53.59\% on average) and non-continuous refinement (52.55\% on average) reveals a meaningful performance gap of over 1\%. It is notable that this gap carries weight, given that each benchmark example involves a single task and prompt-based router updates already constitute a strong baseline. This improvement indicates that routing optimization benefits from curriculum-like approaches that gradually refine the model's expert selection mechanisms, leading to more stable and effective adaptations during test-time optimization.

\section{Analysis and Discussions} 

\subsection{Why Does test-time rerouting Work?}
To understand the mechanisms behind test-time rerouting, we perform a comprehensive analysis of the  DeepSeek-V2-Lite model, examining pathway shifts, expert utilization dynamics, and the evolution of routing confidence.

\begin{figure}[h]
\label{fig:ana}
    \centering
    \includegraphics[width=1.0\textwidth ]{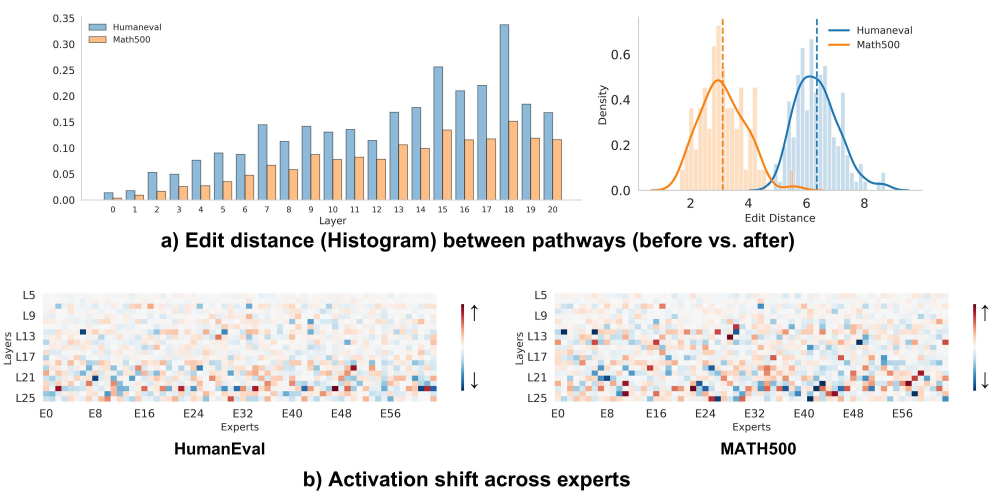}
    \caption{Analysis of test-time rerouting mechanisms across different datasets in DeepSeek-V2-Lite. a) Edit distance before and after rerouting. b) Expert utilization dynamics before and after rerouting.}
    \label{explain}
\end{figure}

\textbf{Rewiring improves Expert Pathways.}
We first examine how expert pathways change after rerouting using edit distance following \citep{li2025find} (described in \cref{pathdiff}), which quantifies pathway differences using Levenshtein edit distance. 
This captures mismatches in expert selection, and pathway shifts. We track the average pairwise edit distance across samples during rerouting. 


The results are displayed in Figure~\ref{explain} a), which shows distinct layer-wise editing patterns. In HumanEval, edit distances peak in deeper layers (18–22, $>$0.35) while earlier ones (5–10) remain minimal ($<$0.05). MATH500 exhibits a similar but stronger trend, with peaks up to 0.16 in layers 20–22, suggesting that deeper layers are more involved in mathematical reasoning. The sample-wise distributions (right panels) further emphasize task-specific differences, indicating high variability in pathway changes across problems. Overall, these results highlight the heterogeneity of expert routing across tasks and demonstrate how our method adaptively reroute pathways.

\textbf{Rewiring highlights Task-Specific Experts.}
We further examine expert-level activation shifts on HumanEval and MATH500 before and after rerouting to understand how the model redistributes computation. Figure~\ref{explain} b) shows heatmaps across layers and experts (red: increased, blue: decreased, normalized for visualization). The results reveal: (1) Adaptive targeting. changes concentrate on a subset of experts, indicating strategic rather than uniform redistribution; (2) Deep-layer adaptation. later layers (L21–L25) undergo stronger shifts, consistent with our edit-distance findings; and (3) Task-specific specialization. HumanEval and MATH500 exhibit distinct patterns, highlighting task-dependent routing behaviors.
This selective activation shows that rerouting strategically emphasizes task-relevant experts, enhancing efficiency by focusing computation on where it is most needed.

\textbf{Rewiring Increases Router Confidence.}
Moreover, we quantify routing confidence by tracking the entropy of expert-selection distributions across all layers during optimization. Lower entropy reflects more focused expert allocation, whereas higher entropy indicates more diffuse routing patterns.

\begin{figure}[h]
    \centering
    \begin{minipage}{0.45\textwidth}
        \centering
        \includegraphics[width=\textwidth]{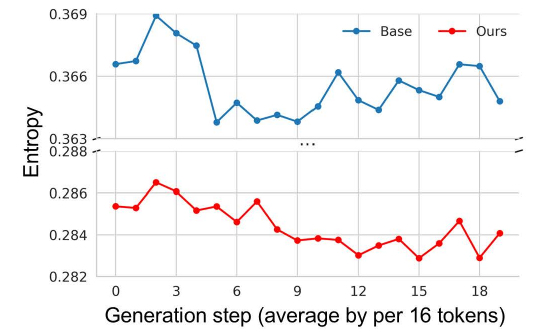}
        \captionof{figure}{Expert routing entropy as a function of sequence length averaged over 16 token blocks.}
        \label{fig:explain}
    \end{minipage}%
    \hfill
    \begin{minipage}{0.5\textwidth}
        \centering
        \small
        \captionof{table}{Computational efficiency comparison across different methods.}
        \begin{tabular}{l|c|c}
        \toprule
        \textbf{Method} & \textbf{Total FLOPs} & \textbf{Time (s)} \\
        \midrule
        Baseline & 4.71e+11 & 10.71s  \\
        ICL(3-shot) & 1.93e+12 & 12.17s \\
        ICL(5-shot) & 3.05e+12 & 12.53s \\
        Self-Consistency (3) &  1.68e+12 & 34.20s \\
        C3PO(100-reference) & 2.63e+12 & 26.20s \\
        \rowcolor{blue!15}
        Rewiring (Ours) & 1.96e+12 & 20.12s \\
        \bottomrule
        \end{tabular}        \label{tab:efficiency}
    \end{minipage}
    \vspace{-.3cm}
\end{figure}

Figure~\ref{fig:explain} shows that our method (red line) exhibits a gradual decrease in router entropy over 18 generation steps, whereas the baseline (blue line) maintains a higher router entropy with pronounced fluctuations. This suggests that our approach progressively develops more focused routing decisions, while the baseline continues to rely on more diffuse expert-selection patterns.
This indicates that our method facilitates concentrating on relevant experts without disrupting established routing mechanisms, thereby enabling more efficient expert utilization for the given tasks.

\subsection{Combining Rewiring with other test-time strategies}
A key advantage of our approach is its plug-and-play compatibility with existing test-time techniques. Since our method only modifies routing decisions without altering the underlying generation process, it can be seamlessly integrated with other approaches like in-context learning and parallel generation methods.

\textbf{Synergy with In-Context Learning.} Large pretrained language models demonstrate strong in-context learning, predicting labels from few demonstration pairs without parameter updates. Empirical evidence demonstrates behavior similar to explicit finetuning~\citep{dai2022can, akyurek2022learning, von2023transformers}. As such, we test combining our method with  with 3-shot demonstrations. As shown in Table~\ref{tab:combination_results}, we observe that this combination yields improvements across multiple tasks, even surpassing our method alone. We hypothesize this synergy occurs because our method provides more effective gradient updates that better leverage the contextual information from demonstration examples, helping extract more meaningful patterns from the limited demonstration data.



\textbf{Unlocking Better Self-Consistency.} Self-consistency \citep{wang2022self} improves reasoning by sampling multiple paths and aggregate together. We combine our method with self-consistency by applying rerouting optimization, then generating multiple reasoning paths with optimized routing decisions. For each sample, we generate 3 reasoning paths. For MMLU, MATH500, and GSM8K, we use majority voting to determine the final accuracy. For code generation tasks (MBPP, HumanEval), we report pass@3 scores. shows substantial improvements with an average 3 percentage point gain over self-consistency alone. We hypothesize that our rerouting framework generates higher-quality reasoning chains by selecting more appropriate experts, and when self-consistency aggregates these improved paths, the voting mechanism amplifies the benefits.

\begin{table}[htbp]
\centering
\small
\setlength{\tabcolsep}{5pt}
\renewcommand{\arraystretch}{0.8}
\caption{Performance comparison of individual Test-Time methods versus combined approaches. Online optimization of routing decisions with our rewiring algorithm can be reliably combined iwth other techniques, such as in-context learning (ICL) or self-consistency.}
\label{tab:combination_results}
\begin{tabular}{l*{6}{c}}
\toprule
\textbf{Method} & \textbf{HumanEval} & \textbf{MBPP} & \textbf{GSM8K} & \textbf{MATH500} & \textbf{MMLU} & \textbf{Average} \\
\midrule
Baseline & 50.60 & 58.37 & 72.10 & 22.60 & 50.77 & 50.89 \\
Rewiring (Ours) & \textbf{54.26} & \textbf{62.65} & \textbf{73.62} & \textbf{25.00} & \textbf{52.40} & \textbf{53.59} \\
\midrule
\multicolumn{7}{c}{\textit{In-Context Learning}} \\
\midrule
ICL (3-shot) & 52.44 & 56.81 & 71.10 & 21.60 & 44.33 & 49.26 \\
\rowcolor{blue!15}
\textbf{ICL + Rewiring} & \textbf{53.05} & \textbf{62.65} & \textbf{76.00} & \textbf{27.00} & \textbf{46.53} & \textbf{53.05} \\
\midrule
\multicolumn{7}{c}{\textit{Self-Consistency}} \\
\midrule
Self-Consistency (3) & 51.02 & 70.04 & 75.28 & 26.20 & 51.87 & 54.88 \\
\rowcolor{blue!15}
\textbf{Self-Consistency (3) + Rewiring} & \textbf{55.08} & \textbf{71.21} & \textbf{77.54} & \textbf{27.40} & \textbf{54.20} & \textbf{57.09} \\
\bottomrule
\end{tabular}
\end{table}

\begin{figure}[htbp]
\centering
\centering
\captionof{table}{Performance of our method versus the baseline on AIME datasets using the \textit{GPT-OSS-20b} model. Optimizing rerouting is especially noticeable at improving model certainty, as shown by improved performance at lower pass@k and improved majority voting, even one challenging benchmarks like AIME.}
\label{tab:aime}
\begin{tabular}{cccccccc}
\toprule
\multirow{2}{*}{\textbf{Data / Method}}  
& \multicolumn{3}{c}{\textbf{Pass@k}} & \multicolumn{3}{c}{\textbf{Maj@k}} & \multirow{2}{*}{\textbf{Average}}\\
\cmidrule(lr){2-4} \cmidrule(lr){5-7}
& \textbf{2} & \textbf{4} & \textbf{8} & \textbf{2} & \textbf{4} & \textbf{8} \\
\midrule
\multicolumn{8}{l}{\textit{AIME25}} \\  
Baseline
& 83.21 & 87.90 & 90.00 & 65.36 & 71.76 & 76.67 & 74.29\\
\rowcolor{blue!15}
\textbf{Rewiring (Ours)}
& \textbf{83.81} & 86.19 & 86.67 & \textbf{67.50} & \textbf{76.67} & \textbf{83.33} & \textbf{75.65}\\
\midrule
\multicolumn{8}{l}{\textit{AIME24}} \\  
Baseline
& 78.57 & 83.76 & 86.67 & 60.60 & 67.48 & 70.00 & 69.58\\
\rowcolor{blue!15}
\textbf{Rewiring (Ours)}
& \textbf{81.67} & \textbf{84.95} & 86.67 & \textbf{65.83} & \textbf{72.57} & \textbf{80.00} &\textbf{73.75} \\
\bottomrule
\end{tabular}
\end{figure}

\subsection{Efficiency Analysis} \label{eff}
Beyond performance improvements, we also compare the computational cost of our method with other online approaches to adapt models. Table~\ref{tab:efficiency} reports the results on the HumanEval task with the DeepSeek model, showing the average total FLOPs and inference time per sample across different methods. 
As shown in Table~\ref{tab:efficiency}, our method achieves notable computational savings over most baselines. Although it requires more computation than the vanilla baseline, it remains substantially more efficient than other test-time techniques, using 1.3× fewer FLOPs than C3PO, 1.6× fewer than ICL (5-shot), and comparable to ICL (3-shot) and Self-Consistency (3). Despite the extra routing operations, our method maintains a competitive inference time of 20.12s, indicating that the overhead of online adaptation is modest while still delivering improved performance. 

\looseness -1 Conceptually, our method requires $n$ additional prefill passes on already generated text every $m$ tokens, which, given the ease of parallelization of prefill is a manageable compute increase. In this work, we implement this optimization in a straightforward manner, but a production implementation could be significantly faster by incorporating the optimization into disaggregated-prefill systems, or timing the MoE rewiring event with low-load timespans, for example when waiting for a user to respond. The actual routing changes are self-contained in routing parameters $\delta$ (shaped as number of experts by number of layers), so that on-device storage is feasible, even for many separate conversations.

\subsection{Extension to Long-Reasoning Tasks}

\looseness -1 To evaluate the Generalizability of our approach to modern reasoning models, we also test GPT-OSS-20B~\citep{agarwal2025gpt}. We evaluate on the AIME benchmark~\citep{maa_aime_problems}, a challenging mathematics competition dataset that requires sophisticated multi-step derivations and mathematical reasoning. This setting allows us to assess whether our online rerouting method can improve performance on the demanding reasoning tasks that represent the current frontier of AI capabilities. 
As shown in Table~\ref{tab:aime}, our method improves performance on both AIME datasets, achieving higher average correctness (75.65\% vs. 74.29\% on AIME25; 73.75\% vs. 69.58\% on AIME24). The primary improvements occur in Maj@k metrics rather than Pass@k, suggesting our routing optimization enhances reasoning consistency rather than solution diversity.

\subsection{Robustness to Context Shifts in Multi-turn Scenarios.}

\begin{figure}
\centering
\includegraphics[width=\textwidth]{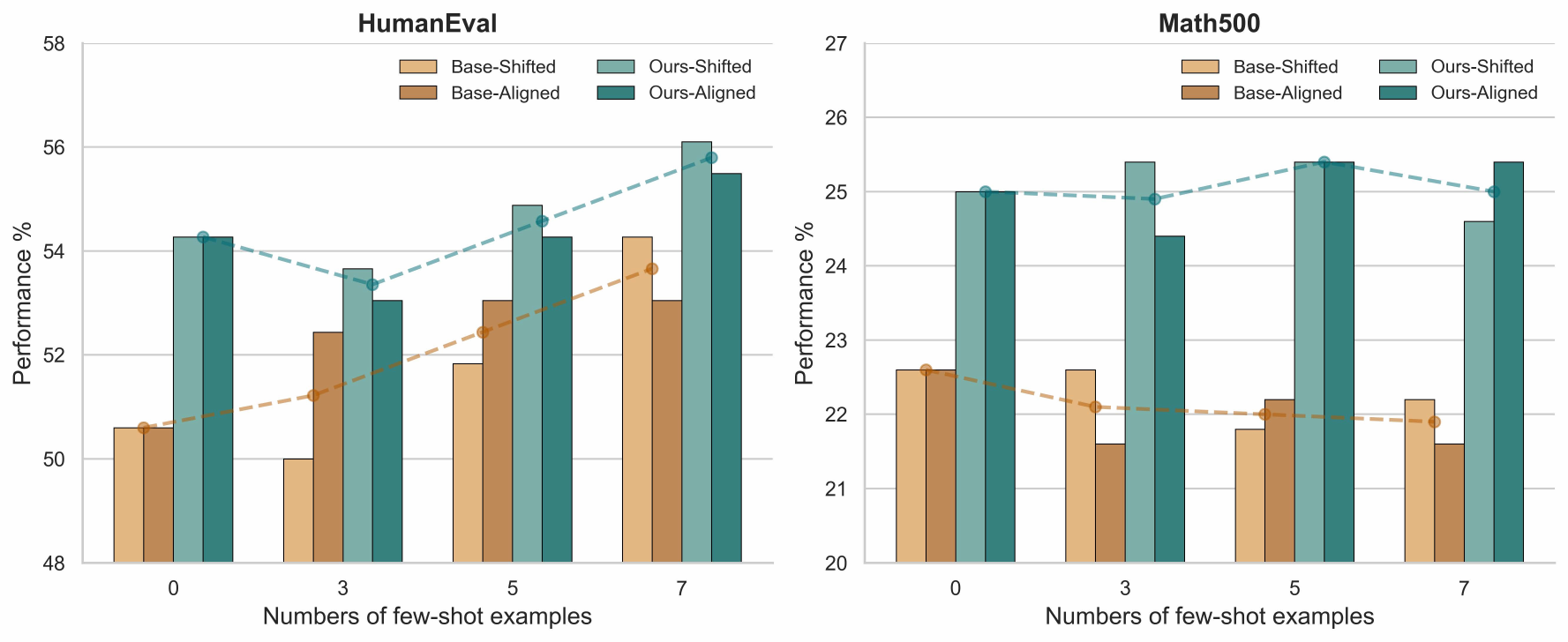}
\caption{Performance of our method versus the baseline across different few-shot examples under shifted and aligned task contexts.\label{fig:context_shift}}
\end{figure}

In real-world applications, MoE models often encounter \textit{multi-turn conversations} where contexts shift dramatically between different topics or tasks. To assess the robustness of our method under realistic multi-turn contexts, we simulate context shifts by prepending few-shot examples from different domains before the target task. Figure~\ref{fig:context_shift} presents results on HumanEval and Math500 using DeepSeek-V2-Lite, under two conditions:
(1) \textbf{Aligned-task}, where few-shot examples are from the same task domain, and
(2) \textbf{Shifted-task}, where examples are drawn from unrelated domains (e.g., MATH and MMLU for code tasks) to induce \textit{cross-domain shifts}.

Across both benchmarks, our method (Ours-Shifted, Ours-Aligned) consistently outperforms the baselines (Base-Shifted, Base-Aligned). On HumanEval, all methods benefit from more-shot examples, but our approach shows a stable improvement. In contrast, baseline gains remain modest, particularly in the Shifted-task setting, where performance fluctuates. On Math500, our method maintains robust and consistent results across both domains, while baselines exhibit limited or even inconsistent improvements.
These findings highlight the robustness and scalability of our test-time rerouting strategy in handling diverse contextual information





\section{Conclusion}
\looseness -1 In this work, we introduce a novel test-time rerouting approach that enables MoE models to dynamically adapt expert selection on the fly, without requiring external data or costly retrieval. The method alternates between routing optimization and steered generation, forming a feedback loop that progressively improves expert selection. To reduce computational overhead, we employ lightweight additive vectors that update only the logits of selected routers.
Extensive experiments show that our approach effectively compensates for the inherent imperfections in MoE routing, yielding consistent gains across multiple benchmarks (up to 6.7\% on code generation) with 1.6× fewer FLOPs than few-shot methods, while maintaining robustness to context shifts. As a plug-and-play regularization strategy, the method flexibly combines with complementary techniques (e.g., Self-Consistency) to further amplify the benefits. More importantly, by introducing a new dimension of plasticity into MoEs, it opens the door to deployment-time adaptation and points toward practical continual self-regulation in MoE models.

\section*{Acknowledgements}
JG acknowledges the support of the Hector II foundation. GS acknowledges the support of the International Max Planck Research School for Intelligent Systems (IMPRS-IS).





\bibliography{iclr2026_conference}

\begin{thebibliography}{43}
\providecommand{\natexlab}[1]{#1}
\providecommand{\url}[1]{\texttt{#1}}
\expandafter\ifx\csname urlstyle\endcsname\relax
  \providecommand{\doi}[1]{doi: #1}\else
  \providecommand{\doi}{doi: \begingroup \urlstyle{rm}\Url}\fi

\bibitem[Agarwal et~al.(2025)Agarwal, Ahmad, Ai, Altman, Applebaum, Arbus, Arora, Bai, Baker, Bao, et~al.]{agarwal2025gpt}
Sandhini Agarwal, Lama Ahmad, Jason Ai, Sam Altman, Andy Applebaum, Edwin Arbus, Rahul~K Arora, Yu~Bai, Bowen Baker, Haiming Bao, et~al.
\newblock gpt-oss-120b \& gpt-oss-20b model card.
\newblock \emph{arXiv preprint arXiv:2508.10925}, 2025.

\bibitem[Aky{\"u}rek et~al.(2022)Aky{\"u}rek, Schuurmans, Andreas, Ma, and Zhou]{akyurek2022learning}
Ekin Aky{\"u}rek, Dale Schuurmans, Jacob Andreas, Tengyu Ma, and Denny Zhou.
\newblock What learning algorithm is in-context learning? investigations with linear models.
\newblock \emph{arXiv preprint arXiv:2211.15661}, 2022.

\bibitem[Aky{\"u}rek et~al.(2024)Aky{\"u}rek, Damani, Qiu, Guo, Kim, and Andreas]{akyurek2024surprising}
Ekin Aky{\"u}rek, Mehul Damani, Linlu Qiu, Han Guo, Yoon Kim, and Jacob Andreas.
\newblock The surprising effectiveness of test-time training for abstract reasoning.
\newblock \emph{arXiv e-prints}, pp.\  arXiv--2411, 2024.

\bibitem[Austin et~al.(2021)Austin, Odena, Nye, Bosma, Michalewski, Dohan, Jiang, Cai, Terry, Le, et~al.]{austin2021program}
Jacob Austin, Augustus Odena, Maxwell Nye, Maarten Bosma, Henryk Michalewski, David Dohan, Ellen Jiang, Carrie Cai, Michael Terry, Quoc Le, et~al.
\newblock Program synthesis with large language models.
\newblock \emph{arXiv preprint arXiv:2108.07732}, 2021.

\bibitem[Brown et~al.(2024)Brown, Juravsky, Ehrlich, Clark, Le, R{\'e}, and Mirhoseini]{brown2024large}
Bradley Brown, Jordan Juravsky, Ryan Ehrlich, Ronald Clark, Quoc~V Le, Christopher R{\'e}, and Azalia Mirhoseini.
\newblock Large language monkeys: Scaling inference compute with repeated sampling.
\newblock \emph{arXiv preprint arXiv:2407.21787}, 2024.

\bibitem[Chen et~al.(2021)Chen, Tworek, Jun, Yuan, Pinto, Kaplan, Edwards, Burda, Joseph, Brockman, et~al.]{chen2021evaluating}
Mark Chen, Jerry Tworek, Heewoo Jun, Qiming Yuan, Henrique Ponde De~Oliveira Pinto, Jared Kaplan, Harri Edwards, Yuri Burda, Nicholas Joseph, Greg Brockman, et~al.
\newblock Evaluating large language models trained on code.
\newblock \emph{arXiv preprint arXiv:2107.03374}, 2021.

\bibitem[Cobbe et~al.(2021)Cobbe, Kosaraju, Bavarian, Chen, Jun, Kaiser, Plappert, Tworek, Hilton, Nakano, et~al.]{cobbe2021training}
Karl Cobbe, Vineet Kosaraju, Mohammad Bavarian, Mark Chen, Heewoo Jun, Lukasz Kaiser, Matthias Plappert, Jerry Tworek, Jacob Hilton, Reiichiro Nakano, et~al.
\newblock Training verifiers to solve math word problems.
\newblock \emph{arXiv preprint arXiv:2110.14168}, 2021.

\bibitem[Dahlke et~al.(2025)Dahlke, Klagges, Zecha, Merkel, Rohr, and Klemm]{dahlke2025mixture}
Robert Dahlke, Henrik Klagges, Dan Zecha, Benjamin Merkel, Sven Rohr, and Fabian Klemm.
\newblock Mixture of tunable experts--behavior modification of deepseek-r1 at inference time.
\newblock \emph{arXiv preprint arXiv:2502.11096}, 2025.

\bibitem[Dai et~al.(2022)Dai, Sun, Dong, Hao, Ma, Sui, and Wei]{dai2022can}
Damai Dai, Yutao Sun, Li~Dong, Yaru Hao, Shuming Ma, Zhifang Sui, and Furu Wei.
\newblock Why can gpt learn in-context? language models implicitly perform gradient descent as meta-optimizers.
\newblock \emph{arXiv preprint arXiv:2212.10559}, 2022.

\bibitem[Dai et~al.(2024)Dai, Deng, Zhao, Xu, Gao, Chen, Li, Zeng, Yu, Wu, et~al.]{dai2024deepseekmoe}
Damai Dai, Chengqi Deng, Chenggang Zhao, RX~Xu, Huazuo Gao, Deli Chen, Jiashi Li, Wangding Zeng, Xingkai Yu, Yu~Wu, et~al.
\newblock Deepseekmoe: Towards ultimate expert specialization in mixture-of-experts language models.
\newblock \emph{arXiv preprint arXiv:2401.06066}, 2024.

\bibitem[Fedus et~al.(2022)Fedus, Zoph, and Shazeer]{fedus2022switch}
William Fedus, Barret Zoph, and Noam Shazeer.
\newblock Switch transformers: Scaling to trillion parameter models with simple and efficient sparsity.
\newblock \emph{Journal of Machine Learning Research}, 23\penalty0 (120):\penalty0 1--39, 2022.

\bibitem[Feng et~al.(2020)Feng, Guo, Tang, Duan, Feng, Gong, Shou, Qin, Liu, Jiang, et~al.]{feng2020codebert}
Zhangyin Feng, Daya Guo, Duyu Tang, Nan Duan, Xiaocheng Feng, Ming Gong, Linjun Shou, Bing Qin, Ting Liu, Daxin Jiang, et~al.
\newblock Codebert: A pre-trained model for programming and natural languages.
\newblock \emph{arXiv preprint arXiv:2002.08155}, 2020.

\bibitem[Gandelsman et~al.(2022)Gandelsman, Sun, Chen, and Efros]{gandelsman2022test}
Yossi Gandelsman, Yu~Sun, Xinlei Chen, and Alexei Efros.
\newblock Test-time training with masked autoencoders.
\newblock \emph{Advances in Neural Information Processing Systems}, 35:\penalty0 29374--29385, 2022.

\bibitem[Gema et~al.(2024)Gema, Leang, Hong, Devoto, Mancino, Saxena, He, Zhao, Du, Madani, Barale, McHardy, Harris, Kaddour, van Krieken, and Minervini]{gema2024mmlu}
Aryo~Pradipta Gema, Joshua Ong~Jun Leang, Giwon Hong, Alessio Devoto, Alberto Carlo~Maria Mancino, Rohit Saxena, Xuanli He, Yu~Zhao, Xiaotang Du, Mohammad Reza~Ghasemi Madani, Claire Barale, Robert McHardy, Joshua Harris, Jean Kaddour, Emile van Krieken, and Pasquale Minervini.
\newblock Are we done with mmlu?, 2024.

\bibitem[Hardt \& Sun()Hardt and Sun]{hardt2305test}
Moritz Hardt and Yu~Sun.
\newblock Test-time training on nearest neighbors for large language models, 2024.
\newblock \emph{URL https://arxiv. org/abs/2305.18466}.

\bibitem[Hendrycks et~al.(2024)Hendrycks, Burns, Kadavath, Arora, Basart, Tang, Song, and Steinhardt]{hendrycks2024measuring}
Dan Hendrycks, Collin Burns, Saurav Kadavath, Akul Arora, Steven Basart, Eric Tang, Dawn Song, and Jacob Steinhardt.
\newblock Measuring mathematical problem solving with the math dataset, 2021.
\newblock \emph{URL https://arxiv. org/abs/2103.03874}, 2, 2024.

\bibitem[H{\"u}botter et~al.(2024)H{\"u}botter, Bongni, Hakimi, and Krause]{hubotter2024efficiently}
Jonas H{\"u}botter, Sascha Bongni, Ido Hakimi, and Andreas Krause.
\newblock Efficiently learning at test-time: Active fine-tuning of llms.
\newblock \emph{arXiv preprint arXiv:2410.08020}, 2024.

\bibitem[Jiang et~al.(2024)Jiang, Sablayrolles, Roux, Mensch, Savary, Bamford, Chaplot, Casas, Hanna, Bressand, et~al.]{jiang2024mixtral}
Albert~Q Jiang, Alexandre Sablayrolles, Antoine Roux, Arthur Mensch, Blanche Savary, Chris Bamford, Devendra~Singh Chaplot, Diego de~las Casas, Emma~Bou Hanna, Florian Bressand, et~al.
\newblock Mixtral of experts.
\newblock \emph{arXiv preprint arXiv:2401.04088}, 2024.

\bibitem[Lepikhin et~al.(2020)Lepikhin, Lee, Xu, Chen, Firat, Huang, Krikun, Shazeer, and Chen]{lepikhin2020gshard}
Dmitry Lepikhin, HyoukJoong Lee, Yuanzhong Xu, Dehao Chen, Orhan Firat, Yanping Huang, Maxim Krikun, Noam Shazeer, and Zhifeng Chen.
\newblock Gshard: Scaling giant models with conditional computation and automatic sharding.
\newblock \emph{arXiv preprint arXiv:2006.16668}, 2020.

\bibitem[Li et~al.(2025{\natexlab{a}})Li, Li, and Zhou]{li2025c3po}
Zhongyang Li, Ziyue Li, and Tianyi Zhou.
\newblock C3po: Critical-layer, core-expert, collaborative pathway optimization for test-time expert re-mixing.
\newblock \emph{arXiv preprint arXiv:2504.07964}, 2025{\natexlab{a}}.

\bibitem[Li et~al.(2025{\natexlab{b}})Li, Fan, and Zhou]{li2025find}
Ziyue Li, Chenrui Fan, and Tianyi Zhou.
\newblock Where to find grokking in llm pretraining? monitor memorization-to-generalization without test.
\newblock \emph{arXiv preprint arXiv:2506.21551}, 2025{\natexlab{b}}.

\bibitem[Lightman et~al.(2023)Lightman, Kosaraju, Burda, Edwards, Baker, Lee, Leike, Schulman, Sutskever, and Cobbe]{lightman2023let}
Hunter Lightman, Vineet Kosaraju, Yuri Burda, Harrison Edwards, Bowen Baker, Teddy Lee, Jan Leike, John Schulman, Ilya Sutskever, and Karl Cobbe.
\newblock Let's verify step by step.
\newblock In \emph{The Twelfth International Conference on Learning Representations}, 2023.

\bibitem[Liu et~al.(2024)Liu, Feng, Wang, Wang, Liu, Zhao, Dengr, Ruan, Dai, Guo, et~al.]{liu2024deepseek}
Aixin Liu, Bei Feng, Bin Wang, Bingxuan Wang, Bo~Liu, Chenggang Zhao, Chengqi Dengr, Chong Ruan, Damai Dai, Daya Guo, et~al.
\newblock Deepseek-v2: A strong, economical, and efficient mixture-of-experts language model.
\newblock \emph{arXiv preprint arXiv:2405.04434}, 2024.

\bibitem[{MAA Committees}()]{maa_aime_problems}
{MAA Committees}.
\newblock {AIME} problems and solutions.
\newblock \url{https://artofproblemsolving.com/wiki/index.php/AIME_Problems_and_Solutions}.

\bibitem[Madaan et~al.(2023)Madaan, Tandon, Gupta, Hallinan, Gao, Wiegreffe, Alon, Dziri, Prabhumoye, Yang, et~al.]{madaan2023self}
Aman Madaan, Niket Tandon, Prakhar Gupta, Skyler Hallinan, Luyu Gao, Sarah Wiegreffe, Uri Alon, Nouha Dziri, Shrimai Prabhumoye, Yiming Yang, et~al.
\newblock Self-refine: Iterative refinement with self-feedback.
\newblock \emph{Advances in Neural Information Processing Systems}, 36:\penalty0 46534--46594, 2023.

\bibitem[Muennighoff et~al.(2024)Muennighoff, Soldaini, Groeneveld, Lo, Morrison, Min, Shi, Walsh, Tafjord, Lambert, et~al.]{muennighoff2024olmoe}
Niklas Muennighoff, Luca Soldaini, Dirk Groeneveld, Kyle Lo, Jacob Morrison, Sewon Min, Weijia Shi, Pete Walsh, Oyvind Tafjord, Nathan Lambert, et~al.
\newblock Olmoe: Open mixture-of-experts language models.
\newblock \emph{arXiv preprint arXiv:2409.02060}, 2024.

\bibitem[Osowiechi et~al.(2023)Osowiechi, Hakim, Noori, Cheraghalikhani, Ben~Ayed, and Desrosiers]{osowiechi2023tttflow}
David Osowiechi, Gustavo A~Vargas Hakim, Mehrdad Noori, Milad Cheraghalikhani, Ismail Ben~Ayed, and Christian Desrosiers.
\newblock Tttflow: Unsupervised test-time training with normalizing flow.
\newblock In \emph{Proceedings of the IEEE/CVF Winter Conference on Applications of Computer Vision}, pp.\  2126--2134, 2023.

\bibitem[Shazeer et~al.(2017)Shazeer, Mirhoseini, Maziarz, Davis, Le, Hinton, and Dean]{shazeer2017outrageously}
Noam Shazeer, Azalia Mirhoseini, Krzysztof Maziarz, Andy Davis, Quoc Le, Geoffrey Hinton, and Jeff Dean.
\newblock Outrageously large neural networks: The sparsely-gated mixture-of-experts layer.
\newblock \emph{arXiv preprint arXiv:1701.06538}, 2017.

\bibitem[Shi et~al.(2024)Shi, Yang, Zhu, Wang, Wu, Li, Cai, Yang, and Meng]{shi2024unchosen}
Chufan Shi, Cheng Yang, Xinyu Zhu, Jiahao Wang, Taiqiang Wu, Siheng Li, Deng Cai, Yujiu Yang, and Yu~Meng.
\newblock Unchosen experts can contribute too: Unleashing moe models’ power by self-contrast.
\newblock \emph{Advances in Neural Information Processing Systems}, 37:\penalty0 136897--136921, 2024.

\bibitem[Sun et~al.(2020)Sun, Wang, Liu, Miller, Efros, and Hardt]{sun2020test}
Yu~Sun, Xiaolong Wang, Zhuang Liu, John Miller, Alexei Efros, and Moritz Hardt.
\newblock Test-time training with self-supervision for generalization under distribution shifts.
\newblock In \emph{International conference on machine learning}, pp.\  9229--9248. PMLR, 2020.

\bibitem[Sun et~al.(2024)Sun, Li, Dalal, Xu, Vikram, Zhang, Dubois, Chen, Wang, Koyejo, et~al.]{sun2024learning}
Yu~Sun, Xinhao Li, Karan Dalal, Jiarui Xu, Arjun Vikram, Genghan Zhang, Yann Dubois, Xinlei Chen, Xiaolong Wang, Sanmi Koyejo, et~al.
\newblock Learning to (learn at test time): Rnns with expressive hidden states.
\newblock \emph{arXiv preprint arXiv:2407.04620}, 2024.

\bibitem[Team(2024)]{team2024qwen2}
Qwen Team.
\newblock Qwen2 technical report.
\newblock \emph{arXiv preprint arXiv:2407.10671}, 2, 2024.

\bibitem[Team(2025)]{team2025qwen3}
Qwen Team.
\newblock Qwen3: Think deeper, act faster, 2025.
\newblock \emph{URL https://qwenlm. github. io/blog/qwen3/. Accessed}, 5\penalty0 (10), 2025.

\bibitem[Von~Oswald et~al.(2023)Von~Oswald, Niklasson, Randazzo, Sacramento, Mordvintsev, Zhmoginov, and Vladymyrov]{von2023transformers}
Johannes Von~Oswald, Eyvind Niklasson, Ettore Randazzo, Jo{\~a}o Sacramento, Alexander Mordvintsev, Andrey Zhmoginov, and Max Vladymyrov.
\newblock Transformers learn in-context by gradient descent.
\newblock In \emph{International Conference on Machine Learning}, pp.\  35151--35174. PMLR, 2023.

\bibitem[Wang et~al.(2020)Wang, Shelhamer, Liu, Olshausen, and Darrell]{wang2020tent}
Dequan Wang, Evan Shelhamer, Shaoteng Liu, Bruno Olshausen, and Trevor Darrell.
\newblock Tent: Fully test-time adaptation by entropy minimization.
\newblock \emph{arXiv preprint arXiv:2006.10726}, 2020.

\bibitem[Wang et~al.(2025)Wang, Chen, Wang, He, Xu, Liang, Liu, Yao, Wang, Ma, et~al.]{wang2025two}
Mengru Wang, Xingyu Chen, Yue Wang, Zhiwei He, Jiahao Xu, Tian Liang, Qiuzhi Liu, Yunzhi Yao, Wenxuan Wang, Ruotian Ma, et~al.
\newblock Two experts are all you need for steering thinking: Reinforcing cognitive effort in moe reasoning models without additional training.
\newblock \emph{arXiv preprint arXiv:2505.14681}, 2025.

\bibitem[Wang et~al.(2022)Wang, Wei, Schuurmans, Le, Chi, Narang, Chowdhery, and Zhou]{wang2022self}
Xuezhi Wang, Jason Wei, Dale Schuurmans, Quoc Le, Ed~Chi, Sharan Narang, Aakanksha Chowdhery, and Denny Zhou.
\newblock Self-consistency improves chain of thought reasoning in language models.
\newblock \emph{arXiv preprint arXiv:2203.11171}, 2022.

\bibitem[Wei et~al.(2022)Wei, Wang, Schuurmans, Bosma, Xia, Chi, Le, Zhou, et~al.]{wei2022chain}
Jason Wei, Xuezhi Wang, Dale Schuurmans, Maarten Bosma, Fei Xia, Ed~Chi, Quoc~V Le, Denny Zhou, et~al.
\newblock Chain-of-thought prompting elicits reasoning in large language models.
\newblock \emph{Advances in neural information processing systems}, 35:\penalty0 24824--24837, 2022.

\bibitem[Welleck et~al.(2024)Welleck, Bertsch, Finlayson, Schoelkopf, Xie, Neubig, Kulikov, and Harchaoui]{welleck2024decoding}
Sean Welleck, Amanda Bertsch, Matthew Finlayson, Hailey Schoelkopf, Alex Xie, Graham Neubig, Ilia Kulikov, and Zaid Harchaoui.
\newblock From decoding to meta-generation: Inference-time algorithms for large language models.
\newblock \emph{arXiv preprint arXiv:2406.16838}, 2024.

\bibitem[Xie et~al.(2024)Xie, Goyal, Zheng, Kan, Lillicrap, Kawaguchi, and Shieh]{xie2024monte}
Yuxi Xie, Anirudh Goyal, Wenyue Zheng, Min-Yen Kan, Timothy~P Lillicrap, Kenji Kawaguchi, and Michael Shieh.
\newblock Monte carlo tree search boosts reasoning via iterative preference learning.
\newblock \emph{arXiv preprint arXiv:2405.00451}, 2024.

\bibitem[Zhou et~al.(2023)Zhou, Yan, Shlapentokh-Rothman, Wang, and Wang]{zhou2023language}
Andy Zhou, Kai Yan, Michal Shlapentokh-Rothman, Haohan Wang, and Yu-Xiong Wang.
\newblock Language agent tree search unifies reasoning acting and planning in language models.
\newblock \emph{arXiv preprint arXiv:2310.04406}, 2023.

\bibitem[Zhou et~al.(2022)Zhou, Lei, Liu, Du, Huang, Zhao, Dai, Le, Laudon, et~al.]{zhou2022mixture}
Yanqi Zhou, Tao Lei, Hanxiao Liu, Nan Du, Yanping Huang, Vincent Zhao, Andrew~M Dai, Quoc~V Le, James Laudon, et~al.
\newblock Mixture-of-experts with expert choice routing.
\newblock \emph{Advances in Neural Information Processing Systems}, 35:\penalty0 7103--7114, 2022.

\bibitem[Zuo et~al.(2025)Zuo, Zhang, Sheng, Qu, Cui, Zhu, Li, Zhang, Long, Hua, et~al.]{zuo2025ttrl}
Yuxin Zuo, Kaiyan Zhang, Li~Sheng, Shang Qu, Ganqu Cui, Xuekai Zhu, Haozhan Li, Yuchen Zhang, Xinwei Long, Ermo Hua, et~al.
\newblock Ttrl: Test-time reinforcement learning.
\newblock \emph{arXiv preprint arXiv:2504.16084}, 2025.

\end{thebibliography}
\bibliographystyle{iclr2026_conference}

\appendix
\section{Experimental Settings} \label{app:a1}
\subsection{Benchmarks}
\textbf{MMLU-redux} is a manually re-annotated subset of the original MMLU benchmark designed to address quality issues in the dataset. The dataset contains 3,000 questions across 30 MMLU subjects (100 questions per subject), with expert annotators identifying and categorizing various types of errors using a comprehensive error taxonomy. These errors include issues such as bad question clarity, unclear options, no correct answers, multiple correct answers, and wrong ground truth labels. MMLU-Redux provides a more reliable evaluation standard by filtering out problematic questions and offering corrected annotations where possible.

\textbf{HumanEval} is a benchmark dataset for evaluating code generation capabilities of large language models. The dataset consists of 164 hand-crafted programming problems, each including a function signature, docstring, body, and several unit tests (averaging 7.7 tests per problem). These challenges assess language comprehension, algorithms, and simple mathematics, with difficulty comparable to simple software interview questions.

\textbf{MBPP} is a code generation benchmark consisting of around 1,000 crowd-sourced Python programming problems designed to be solvable by entry-level programmers. Each problem includes a task description, code solution, and 3 automated test cases, covering programming fundamentals and standard library functionality. The dataset provides two versions: a full version with 974 problems and a sanitized version with 427 problems. The sanitized split underwent a second round of annotation to improve task descriptions, addressing issues where the original descriptions might not be sufficiently expressive to solve the tasks. This hand-verified subset provides higher-quality problem statements for more reliable evaluation of code generation models. In this paper, we use the sanitized version.

\textbf{GSM8K} is a dataset of 8,500 high-quality, linguistically diverse grade school math word problems created by human problem writers. The dataset is segmented into 7,473 training problems and 1,319 test problems. Each problem takes between 2 and 8 steps to solve using basic arithmetic operations, with problems designed so that a bright middle school student should be able to solve every problem. Solutions are provided in natural language format rather than pure mathematical expressions, offering insight into multi-step reasoning processes.

\textbf{MATH-500} is a curated subset of 500 challenging mathematical problems selected from the MATH dataset \citep{hendrycks2024measuring}. These problems span seven topics: Prealgebra, Algebra, Number Theory, Counting and Probability, Geometry, Intermediate Algebra, and Precalculus. Each problem requires multi-step reasoning and is designed to test a model's ability to apply mathematical principles, execute complex calculations, and communicate solutions clearly.

\subsection{Baselines}
\textbf{C3PO}
We adopt C3PO (Critical-Layer, Core-Expert, Collaborative Pathway Optimization)~\citep{li2025c3po} as our primary baseline method. C3PO is a test-time optimization approach designed to address the suboptimal expert pathway selection problem in Mixture of Experts (MoE) large language models. The method operates on the observation that end-to-end trained routers often produce inefficient pathways for challenging or out-of-distribution samples, leading to degraded performance on diverse downstream tasks.

The core idea of C3PO is to dynamically re-mix expert pathways during inference by leveraging successful routing patterns from a reference set. Given a reference set of $m$ samples $\{(x_i, y_i)\}_{i=1}^m$ with their corresponding expert pathway matrices $\{\omega_i\}_{i=1}^m$ (where each $\omega_i \in \mathbb{R}^{L \times E}$ with $L$ layers and $E$ experts) on which the model makes correct predictions, C3PO aims to find an improved pathway matrix $\omega$ for a new test sample $x$ that leads to more accurate outputs.

Among the three optimization strategies proposed by C3PO (gradient descent, kernel regression, and mode finding), we implement the \textbf{kernel regression} approach in our experiments. This method estimates optimal expert pathways by computing a weighted average of neighbors' pathway matrices:

\begin{equation}
\hat{\omega} = \frac{\sum_{i \in \mathcal{N}(x)} K(x_i, x) \omega_i}{\sum_{i \in \mathcal{N}(x)} K(x_i, x)}
\end{equation}

where $K(\cdot, \cdot)$ is a kernel function measuring sample similarity, and $\mathcal{N}(x)$ denotes the neighborhood of $x$ in the reference set. The final pathway is obtained through interpolation:

\begin{equation}
\omega \leftarrow \alpha \omega + (1 - \alpha) \hat{\omega}
\end{equation}

where $\alpha$ is optimally chosen to minimize the loss function.

For our comparative evaluation, we construct reference sets for each benchmark as follows:

For HumanEval and MBPP, we use the MBPP validation set; for GSM8K and MATH500, we utilize the GSM8K training set; and for MMLU, we employ the MMLU training set. From each source, we randomly sample 100 instances where the model produces the correct predictions, ensuring high-quality pathway references for the optimization process.

\textbf{In-Context Learning } In-Context Learning (ICL) is a fundamental capability of large language models that enables them to adapt to new tasks by using a few demonstration examples provided in the input prompt, without requiring parameter updates~\citep{wang2022self,wei2022chain}. 

In our experimental setup, we implement ICL as a baseline across all evaluation benchmarks. The few-shot examples are carefully selected from relevant datasets to ensure domain alignment and high-quality demonstrations:

\begin{itemize}
\item \textbf{HumanEval} and \textbf{MBPP}: We sample 3-5 examples from the MBPP prompt collection, which provides well-crafted code generation examples with clear problem descriptions and corresponding Python solutions.

\item \textbf{GSM8K}: We utilize 3-5 examples from the GSM8K training set, featuring step-by-step mathematical reasoning demonstrations that guide the model through problem-solving processes.

\item \textbf{MATH500}: We sample 3-5 examples from the HYDRA-Math dataset, which offers high-quality mathematical problem-solution pairs covering various difficulty levels and mathematical domains.

\item \textbf{MMLU}: We select 3-5 examples from the MMLU validation set, ensuring coverage of diverse knowledge domains while maintaining format consistency for multiple-choice questions.
\end{itemize}

For each benchmark, the few-shot examples are randomly sampled from their respective source datasets and prepended to each test query. This approach provides the model with task-specific context while maintaining consistency across different evaluation scenarios. The number of examples (3-5) is chosen to balance between providing sufficient context and avoiding excessive prompt length that might degrade model performance.

\subsection{Model Selection}
\textbf{OLMoE} is a fully open-source MoE language model developed by the Allen Institute for AI and Contextual AI \citep{muennighoff2024olmoe}. The model employs a decoder-only transformer architecture with 1 billion active and 7 billion total parameters. Each MoE layer contains 64 experts, of which 8 are activated per input token through a learned router network. This sparse activation mechanism enables computational efficiency similar to dense 1B parameter models while leveraging the full 7B parameter capacity.

\textbf{DeepSeek-V2-Lite} is a smaller variant of the DeepSeek-V2 model family, developed by DeepSeek-AI~\citep{liu2024deepseek}. The model employs innovative architectures, including Multi-head Latent Attention (MLA) and DeepSeekMoE, with 15.7 billion total parameters and 2.4 billion activated parameters per token. DeepSeek-V2-Lite has 27 layers with a hidden dimension of 2048, utilizing MLA with 16 attention heads, where each head has a dimension of 128.
The model adopts the DeepSeekMoE architecture, where all feed-forward networks except the first layer are replaced with MoE layers. Each MoE layer consists of 2 shared experts and 64 routed experts, with 6 experts activated for each token. This sparse activation mechanism enables computational efficiency while maintaining strong performance across diverse tasks.

\textbf{Qwen1.5-MoE} is developed by the Qwen team~\citep{team2024qwen2}. The model is upcycled from the dense Qwen-1.8B model, featuring 14.3 billion total parameters with 2.7 billion activated parameters during runtime. Despite using only 2.7B active parameters, the model achieves comparable performance to Qwen1.5-7B while requiring 75\% fewer training resources and demonstrating 1.74x faster inference speed.
The model employs a fine-grained expert architecture with 64 experts total, consisting of 4 shared experts that are always activated and 60 routing experts with 4 activated per token. This configuration represents an 8-fold increase in expert count compared to conventional MoE setups, enabling higher model capacity without proportional parameter increases. The fine-grained expert design partitions a single FFN into multiple segments, each serving as an individual expert, allowing for more specialized knowledge representation.

\textbf{GPTOSS-20B} is OpenAI's recently released open-source MoE model with 21B total parameters but only ~3.6B active at inference \citep{agarwal2025gpt}. Built on a Mixture-of-Experts architecture with 24 layers and 32 experts using Top-4 routing, it represents a state-of-the-art reasoning model that achieves performance similar to OpenAI o3-mini on reasoning benchmarks. Its sophisticated expert routing system and focus on mathematical reasoning make it an ideal candidate for evaluating our test-time rerouting method on challenging tasks like AIME.

\section{Method Details} 
\subsection{Test-Time MoE Rerouting}
We elaborate the rerouting algorithmic pipeline in \cref{alg:tamoe}, which details the step-by-step procedure for adaptive expert selection and pathway calibration at test time.
\subsection{Pathway Differences} \label{pathdiff}
We first examine how expert pathways change after rerouting using edit distance following \citep{li2025find}. For input $x_i$, we define its pathway $s_i$ as the ordered sequence of selected experts across $L$ MoE layers as $s_i = \text{concat}(e_1^{(i)}, e_2^{(i)}, \ldots, e_L^{(i)})$, where $e_\ell^{(i)}$ represents expert indices at layer $\ell$ as comma-separated strings (e.g., `3,1,5'), joined across layers with hyphens.
We quantify pathway differences using Levenshtein edit distance as $D_{\text{path}}(s_i, s_j) = \text{EditDistance}(s_i, s_j)$. This captures mismatches in expert selection, and pathway shifts. 

\section{Additional Analysis} 
\subsection{Layer-wise confidence distributions across different tasks} 
\begin{figure}[h]
    \centering
    \includegraphics[width=1.0\textwidth ]{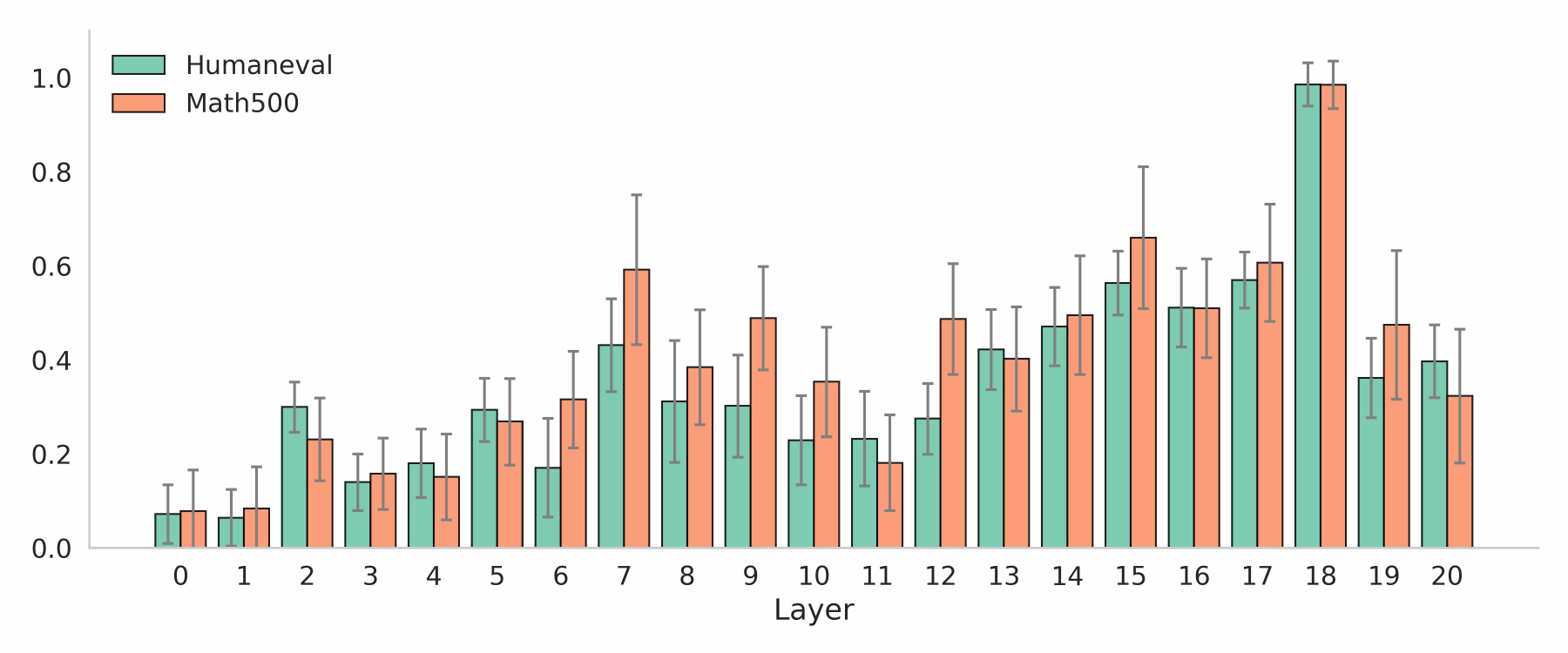}
    \caption{Layer-wise confidence distributions across different tasks in DeepSeek-V2-lite-MoE}
    \label{fig:layers}

\end{figure}

We visualize the confidence distributions across different tasks using DeepSeek-V2-lite as an example. As illustrated in Figure~\ref{fig:layers}, activation patterns across experts and layers are highly task-specific. For instance, math tasks (Math500) and code tasks (HumanEval) exhibit distinct confidence patterns across layers, with math tasks showing higher confidence in middle layers (layers 7-14) while code tasks demonstrate more distributed confidence patterns with peaks in later layers (layers 15-18). Notably, both tasks show generally higher confidence concentrated in the middle-to-later layers compared to early layers.

\subsection{Sensitive Analyses of Hyper-parameters}

\begin{figure}[h]
\label{fig:ana}
    \centering
    \includegraphics[width=1.0\textwidth ]{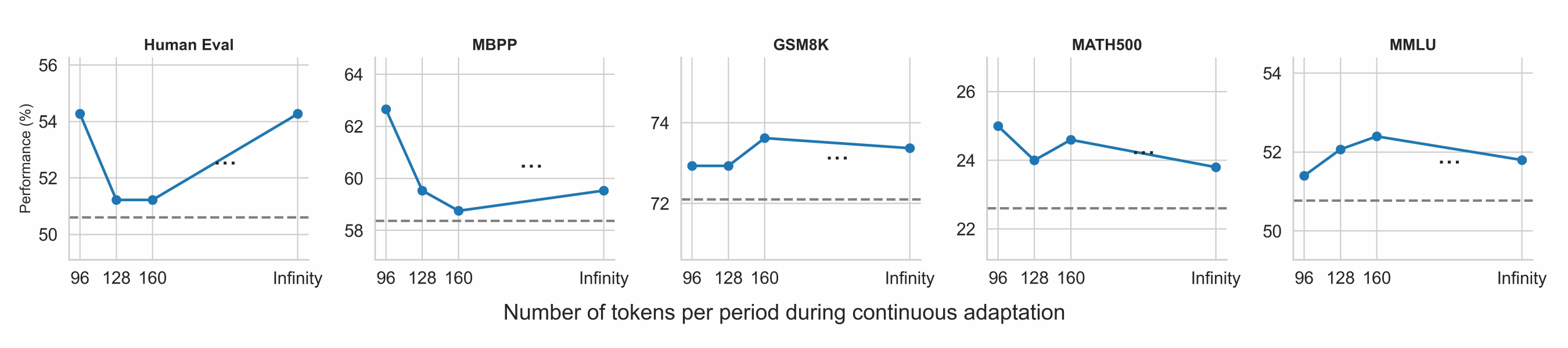}
    \caption{Effect of optimization interval on performance across 5 benchmarks. The dashed line denotes the baseline results without rerouting.}
    \label{fig:sen}
    
\end{figure}

As shown in Figure~\ref{fig:sen}, the optimization interval analysis across five benchmarks shows that our test-time rerouting consistently outperforms baseline performance. Intervals of 128-160 tokens achieve optimal performance, balancing routing adaptability with computational efficiency. Both frequent re-optimization (96 tokens) and no updates degrade performance due to over-adaptation and loss of dynamic adjustment benefits, respectively. Mathematical reasoning tasks (GSM8K, MATH500) benefit most from proper interval tuning.

\begin{algorithm}[h]
\caption{Test-Time MoE Rerouting}
\label{alg:tamoe}
\textbf{Input:} Pre-trained MoE model $\mathcal{M}$; input prompt $\mathbf{x} = (x_1, \ldots, x_n)$; optimization steps $n$; generation interval $m$; optimizer $\mathcal{O}$.\\
\textbf{Output:} Generated response $\mathbf{y}$.\\

// \textbf{Parameter Initialization}\\
Initialize $\boldsymbol{\delta}^{(l)} = \mathbf{0} \in \mathbb{R}^{N}$ for all layers $l \in \mathcal{L}$;\\

// \textbf{Phase 1: In-Context Routing Optimization}\\
$\mathbf{x}_{\text{current}} = \mathbf{x}$, $T = |\mathbf{x}|$;\\
Select layers $\mathcal{S} = \text{TopK}(\{C^{(l)}\}_{l=1}^L, r)$ or compute soft weights $\{w_l\}_{l=1}^L$;\\
\textbf{for} $i = 1$ to $n$ \textbf{do}\\
\quad Compute loss $\mathcal{L}(\{\boldsymbol{\delta}^{(l)}\}_{l=1}^L) = -\sum_{j=1}^{T-1} \log p(x_{j+1} \mid x_{1:j}, \{\boldsymbol{\delta}^{(l)}\}_{l=1}^L)$;\\
\quad \textbf{if} hard selection \textbf{then}\\
\quad \quad \textbf{for} $l \in \mathcal{S}$ \textbf{do}\\
\quad \quad \quad $\boldsymbol{\delta}^{(l)} \leftarrow \mathcal{O}(\boldsymbol{\delta}^{(l)}, \nabla_{\boldsymbol{\delta}^{(l)}} \mathcal{L})$;\\
\quad \quad \textbf{end for}\\
\quad \textbf{else} // soft weighting\\
\quad \quad \textbf{for} $l = 1$ to $L$ \textbf{do}\\
\quad \quad \quad $\boldsymbol{\delta}^{(l)} \leftarrow \mathcal{O}(\boldsymbol{\delta}^{(l)}, w_l \nabla_{\boldsymbol{\delta}^{(l)}} \mathcal{L})$;\\
\quad \quad \textbf{end for}\\
\quad \textbf{end if}\\
\textbf{end for}\\

// \textbf{Phase 2: Steered Generation with Periodic Re-optimization}\\
Initialize $\mathbf{y} = ()$;\\
\textbf{repeat}\\
\quad // \textit{Generate m tokens using optimized routing parameters}\\
\quad \textbf{for} $k = 1$ to $m$ \textbf{do}\\
\quad \quad Generate $x_{\text{next}} \sim p(\cdot \mid \mathbf{x}_{\text{current}}, \{\boldsymbol{\delta}^{(l)}\}_{l=1}^L)$;\\
\quad \quad Append $x_{\text{next}}$ to $\mathbf{y}$ and $\mathbf{x}_{\text{current}}$;\\
\quad \textbf{end for}\\
\quad // \textit{Re-optimize with extended context}\\
\quad $T = |\mathbf{x}_{\text{current}}|$;\\
\quad Select layers $\mathcal{S} = \text{TopK}(\{C^{(l)}\}_{l=1}^L, r)$ or compute soft weights $\{w_l\}_{l=1}^L$;\\
\quad \textbf{for} $i = 1$ to $n$ \textbf{do}\\
\quad \quad Compute loss $\mathcal{L}(\{\boldsymbol{\delta}^{(l)}\}_{l=1}^L) = -\sum_{j=1}^{T-1} \log p(x_{\text{current},j+1} \mid x_{\text{current},1:j}, \{\boldsymbol{\delta}^{(l)}\}_{l=1}^L)$;\\
\quad \quad \textbf{if} hard selection \textbf{then}\\
\quad \quad \quad \textbf{for} $l \in \mathcal{S}$ \textbf{do}\\
\quad \quad \quad \quad $\boldsymbol{\delta}^{(l)} \leftarrow \mathcal{O}(\boldsymbol{\delta}^{(l)}, \nabla_{\boldsymbol{\delta}^{(l)}} \mathcal{L})$;\\
\quad \quad \quad \textbf{end for}\\
\quad \quad \textbf{else} // soft weighting\\
\quad \quad \quad \textbf{for} $l = 1$ to $L$ \textbf{do}\\
\quad \quad \quad \quad $\boldsymbol{\delta}^{(l)} \leftarrow \mathcal{O}(\boldsymbol{\delta}^{(l)}, w_l \nabla_{\boldsymbol{\delta}^{(l)}} \mathcal{L})$;\\
\quad \quad \quad \textbf{end for}\\
\quad \quad \textbf{end if}\\
\quad \textbf{end for}\\
\textbf{until} end-of-sequence or max length reached;\\

\textbf{return} $\mathbf{y}$;
\end{algorithm}

\subsection{Impact on Sample Diversity in Parallel Generation.}

\begin{table}[H]
    \centering
    \captionof{table}{Diversity evaluation results comparison}
    \begin{tabular}{@{}lcc@{}}
    \toprule
    \textbf{Metric} & \textbf{Baseline} & \textbf{Ours} \\
    \midrule
    Cosine Div. & $0.390 \pm 0.157$ & $0.380 \pm 0.150$ \\
    Semantic Div. & $0.012 \pm 0.007$ & $0.012 \pm 0.008$ \\
    \bottomrule
    \end{tabular}
    \label{tab:diversity_results}
\end{table}

\textbf{Impact on Sample Diversity in Parallel Generation.} While our online adaptation method improves routing efficiency, we investigate whether the adapted routing decisions might reduce sample diversity when generating multiple sequences in parallel. 
To quantify this effect, we evaluate the diversity of generated samples using two complementary metrics: semantic diversity based on CodeBERT embeddings~\citep{feng2020codebert}, which captures functional similarities between code snippets, and TF-IDF-based cosine diversity, which measures lexical variation.
We used the DeepSeek MoE model with default settings to generate 10 samples on the HumanEval task. 
Table~\ref{tab:diversity_results} shows that our method maintains comparable diversity to the baseline across both metrics. Cosine diversity scores are nearly identical ($0.380 \pm 0.150$ vs. $0.390 \pm 0.157$), as are semantic diversity scores ($0.012 \pm 0.008$ vs. $0.012 \pm 0.007$). These results demonstrate that our online adaptation preserves sample diversity while improving routing efficiency.

\end{document}